%% file: egpaper_final.tex
\documentclass[10pt,twocolumn,letterpaper]{article}

\usepackage{iccv}
\usepackage{times}
\usepackage{epsfig}
\usepackage{graphicx}
\usepackage{amsmath}
\usepackage{amssymb}
\usepackage{amsthm}
\usepackage{booktabs}
\usepackage{bm}
\usepackage{caption}
\usepackage{color}
\usepackage{multirow}
\usepackage[ruled,vlined]{algorithm2e}
\usepackage{url}
\usepackage[numbers,sort,compress]{natbib}
\newcommand{\myparagraph}[1]{\vspace{0.2em}\noindent\textbf{#1}}

\definecolor{green}{rgb}{0.55, 0.71, 0.0}
\definecolor{amaranth}{rgb}{0.9, 0.17, 0.31}
\definecolor{amber}{rgb}{1.0, 0.49, 0.0}
\definecolor{azure}{rgb}{0.0, 0.5, 1.0}
\definecolor{byzantine}{rgb}{0.74, 0.2, 0.64}
\definecolor{forestGreen}{rgb}{0.0, 0.54, 0.0}

\usepackage[pagebackref=true,breaklinks=true,letterpaper=true,colorlinks,bookmarks=false]{hyperref}
\usepackage[accsupp]{axessibility}  

\iccvfinalcopy 

\def\iccvPaperID{4338} 

\ificcvfinal\fi

\begin{document}

\title{Learning Motion Priors for 4D Human Body Capture in 3D Scenes}


\newcommand*{\affaddr}[1]{#1} 
\newcommand*{\affmark}[1][*]{\textsuperscript{#1}}
\newcommand*{\email}[1]{\small{\texttt{#1}}}

\author{
Siwei Zhang\affmark[1] \quad
Yan Zhang\affmark[1] \quad
Federica Bogo\affmark[2]  \quad
Marc Pollefeys\affmark[1,2]  \quad
Siyu Tang\affmark[1]\\
\affaddr{\affmark[1]ETH Z\"urich} \quad \affaddr{\affmark[2]Microsoft}\\
\email{\{siwei.zhang, yan.zhang, marc.pollefeys, siyu.tang\}@inf.ethz.ch} \quad \email{febogo@microsoft.com}
}

\twocolumn[{%
\renewcommand\twocolumn[1][]{#1}%
\maketitle
\begin{center}
    \newcommand{\teaserwidth}{\textwidth}
\vspace{-0.2in}
    \centerline{
    \includegraphics[width=1\linewidth]{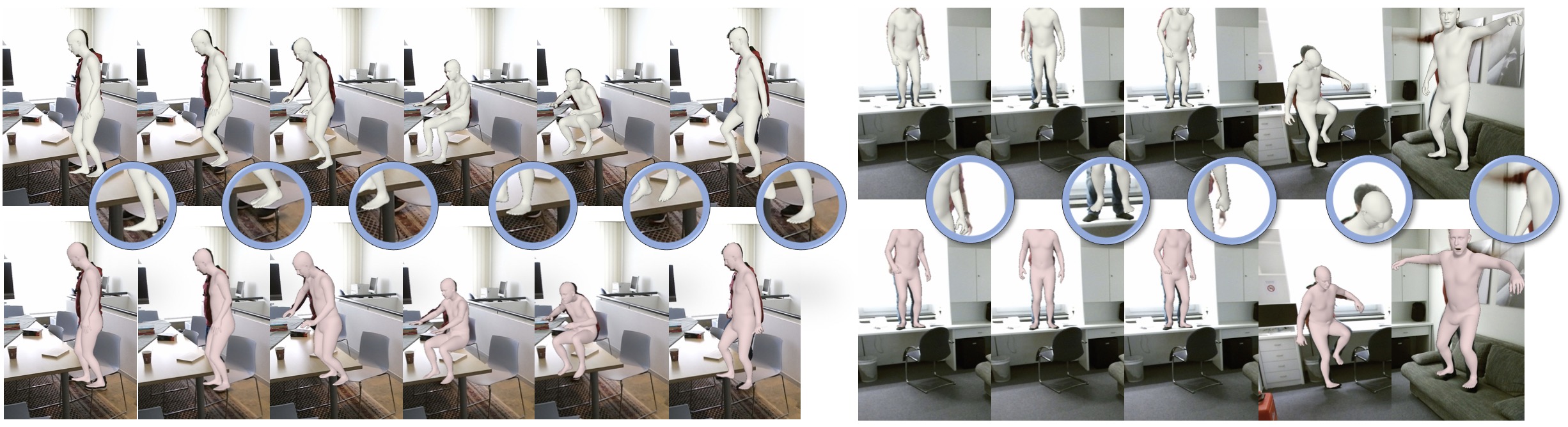}
     }
  \vspace{-0.1in}
  \captionof{figure}{
  By leveraging data-driven motion priors learned from the large-scale mocap dataset AMASS~\cite{mahmood2019amass}, we reconstruct high-quality human motions in complex 3D scenes from monocular RGB(D) input. 
  Our proposed method (second row) robustly deals with occlusions and achieves more accurate motion reconstructions compared with
  PROX~\cite{hassan2019resolving} (first row).
  }

\label{fig:teaser}
\end{center}%
}]

\maketitle
\ificcvfinal\thispagestyle{empty}\fi

\begin{abstract}
   Recovering high-quality 3D human motion in complex scenes from monocular videos is important for many applications, ranging from AR/VR to robotics. However, capturing realistic human-scene interactions, while dealing with occlusions and partial views, is challenging; current approaches are still far from achieving compelling results. 
   We address this problem by proposing LEMO: LEarning human MOtion priors for 4D human body capture.
   By leveraging the large-scale motion capture dataset AMASS~\cite{mahmood2019amass}, we introduce a novel motion smoothness prior, which strongly reduces the jitters exhibited by poses recovered over a sequence. Furthermore, to handle contacts and occlusions occurring frequently in body-scene interactions, we design a contact friction term and a contact-aware motion infiller obtained via per-instance self-supervised training. 
   To prove the effectiveness of the proposed motion priors, we combine them into a novel pipeline for 4D human body capture in 3D scenes.
   With our pipeline, we demonstrate high-quality 4D human body capture, reconstructing smooth motions and physically plausible body-scene interactions. 
   The code and data are available at \url{https://sanweiliti.github.io/LEMO/LEMO.html}.
\end{abstract}

\input{sec1-intro}

\input{sec2-relatedwork}

\input{sec3-method}

\input{sec4-experiment}

\input{sec5-conclusion}

{\small
\bibliographystyle{ieee_fullname}
\bibliography{egbib}
}

\input{sec6-appendix}

\end{document}

%% file: sec1-intro.tex
\section{Introduction}
Recovering realistic human motions in everyday 3D scenes is essential for human behaviour understanding, human-scene interaction synthesis, and virtual avatar creation.  
Marker-based optical motion capture systems (mocap) have the proven capability of recovering highly accurate human motions.  
However, such systems require expert knowledge and expensive setup, making it impractical to capture people in their everyday environments, \eg~recording people in their living rooms, offices or kitchens. 

Recently, PROX~\cite{hassan2019resolving} has been proposed as a lightweight pipeline to capture everyday person-scene interactions from monocular sequences given pre-scanned 3D scene geometries. 
With affordable commodity sensors like an RGB or RGBD camera, it is quite easy to scan a scene and record how humans move in and interact with it.
This shows a promising setup for capturing large-scale human motions in everyday environments. However, as shown in this work\footnote{please see videos on the project page}, the recovered human motions exhibit severe skating and jitters.
The reconstruction quality is far behind that obtained with commercial mocap systems. 
Building a multi-view setup or using additional wearable sensors (\eg~Inertial Measurement Unit (IMUs)) can help improve motion reconstruction quality. 
However, most multi-view settings require careful calibration and synchronization in a controlled environment and IMUs suffer from heading drift and interference. Furthermore, human motions obtained by IMUs~\cite{von2018recovering} or multi-view setups~\cite{joo2015panoptic} still exhibit jitters and are less compelling than the ones from mocap systems.

To improve the naturalness and accuracy of human motions reconstructed from monocular RGBD sequences (\eg~the PROX pipeline~\cite{hassan2019resolving}) and to close the performance gap between the monocular RGBD setup and marker-based mocap systems, we argue that it is essential to leverage data-driven approaches and learn powerful motion priors from high-quality large-scale mocap data (\eg.~AMASS~\cite{mahmood2019amass}). 
To this end, we propose LEMO (\textbf{LE}arning human \textbf{MO}tion priors), which has two key innovations: a marker-based motion smoothness prior and a contact-aware motion infiller which is fine-tuned per-instance in a self-supervised fashion. As shown in the experiments, 
LEMO can effectively capture the intrinsic properties of human motions and regularize the noisy and partial observations. 
As a result, the reconstructed human motions are smooth, physically plausible and robust to occlusions which are inevitable when capturing human motions in everyday 3D scenes. 

\textbf{Marker-based motion smoothness prior.} 
3D human bodies reconstructed by PROX~\cite{hassan2019resolving} have severe jitters over time. Although some heuristic methods like penalizing joint velocity/acceleration can improve temporal smoothness, they also degrade the motion naturalness. As shown in our experiments, they can introduce foot-ground skating artifacts, and may result in invalid body configurations like joint hyperextension.
To capture the holistic full-body dynamics,  we use a fully convolutional autoencoder to aggregate local motion cues in a bottom-up manner, and derive latent motion patterns that cover a large spatio-temporal receptive field. Then, we design a motion smoothness constraint which works in this latent space rather than directly on the body.
To incorporate body shape information and model important degrees-of-freedom (DoFs), \eg~rotation about limb axes, as in~\cite{zhang2020we} we represent the body in each frame by surface markers instead of body joints. 
We learn this convolutional motion smoothness prior on the AMASS~\cite{mahmood2019amass} dataset.
As shown in our experiments, the proposed prior not only significantly increases the reconstruction quality on the PROX dataset, but also improves the motion naturalness on the IMU-based 3DPW dataset \cite{von2018recovering}, suggesting its effectiveness and potential usage for other motion capture and reconstruction settings.

\textbf{Contact-aware motion infiller via per-instance self-supervised learning.}
When capturing humans moving in and interacting with everyday 3D environments (\eg~living rooms or offices), partial body occlusions are almost inevitable. They pose a challenge for reconstruction algorithms, causing invalid poses and foot-ground skating artifacts. 
By leveraging AMASS~\cite{mahmood2019amass}, we learn a neural motion infiller that is able to infer plausible motions of occluded body parts given partial observations. Our network is inspired by~\cite{kaufmann2020infilling}, but goes beyond the previous work to jointly predict the foot contact status and body motion.
%
%
Combined with {\bf a contact friction term} motivated by intuitive physics, the infilled motion is natural, realistic and has proper foot-ground interaction, eliminating the foot skating artifacts.
Furthermore, inspired by~\cite{joo2020exemplar}, we propose a per-instance network fine-tuning scheme. For a test instance which contains partial observations (\eg only the upper body motion as the lower body parts are occluded by the sofa in a 3D scene), we fine-tune the pre-trained motion infilling network by minimizing a self-supervised loss that is defined on the visible body parts.
In this way, we effectively adapt the general motion infilling ``prior'' to per-test-instance, achieving notable improvements both for AMASS~\cite{mahmood2019amass} and PROX~\cite{hassan2019resolving}.

We further carefully combine the learned motion priors and the 
contact friction term into a novel multi-stage optimization pipeline for 4D human body capture in 3D scenes.

\myparagraph{Contributions.} In summary, our contributions are 1) a novel marker-based motion smoothness prior that encodes the ``whole-body'' motion in a learned latent space, which can be easily plugged into an optimization pipeline; 2) a novel contact-aware motion infiller that can be adapted to per-test-instance via self-supervised learning; 3) a new optimization pipeline that explores both learned motion priors and the physics-inspired contact friction term for scene-aware human motion capture. We extensively evaluate the proposed priors and the optimization pipeline. The results show both the wide applicability of the learned motion priors and the efficacy of the optimization pipeline for monocular RGBD human motion capture in 3D scenes.

%% file: sec2-relatedwork.tex
\section{Related Work}
\myparagraph{Human motion recovery from RGB(D) sequences.}
Human motion recovery extends the problem of reconstructing per-frame body 3D shape and pose~\cite{agarwal2005recovering, bualan2008naked, grauman2003inferring, hassan2019resolving,kanazawa2018end,kolotouros2019learning,pavlakos2019expressive, guler2019holopose, omran2018neural, tan2017indirect, tung2017self, xu2019denserac, Sigal:AMDO:2006} to sequences of frames, requiring temporal consistency between estimates.
A number of works tackle the problem adopting skeleton/joint-based representations for the body~\cite{cai2019exploiting,dabral2018learning,liu2020attention, mehta2017vnect, pavllo20193d,rayat2018exploiting,xu2020deep,mehta2018single, mehta2020xnect, zhou2016sparseness, sigal2012loose, elhayek2015efficient, burenius20133d, elhayek2015outdoor, wang2020motion, zhang20204d, kiciroglu2020activemocap}. Working with 3D joints instead of surfaces, these representations cannot adequately model the 3D shape of the body and body-scene interactions. Other works propose to use parametric 3D human models (\eg~SMPL~\cite{loper2015smpl}) to obtain complete 3D body meshes from multi-view~\cite{dong2020motion, huang2017towards, gall2010optimization, joo2018total, saini2019markerless, wang2017outdoor} or monocular RGB(D) sequences~\cite{kanazawa2019learning, kocabas2020vibe, choi2020beyond, sun2019human, luo20203d, zanfir2020weakly, liu20204d}.
Kanazawa et al.~\cite{kanazawa2019learning} learn a temporal context representation to predict motion in past and future frames.
Kocabas et al.~\cite{kocabas2020vibe} use a bi-directional gated recurrent unit (GRU) to temporally encode per-frame image features, and couple it with an adversarial discriminator to distinguish between real and predicted motions.
Choi et al.~\cite{choi2020beyond} propose to better integrate past and future frames' temporal information to increase temporal consistency. 
Sun et al.~\cite{sun2019human} introduce a multi-level framework to decouple body skeleton and more detailed shape and pose information.
Luo et al.~\cite{luo20203d} propose a two-step encoding scheme, which first captures the coarse overall motion by a pretrained motion representation, and then refines these estimates.
Nevertheless, these methods focus only on human body motion reconstruction, ignoring person-scene interactions. 

\myparagraph{Person-scene interaction.}
Hasler et al.~\cite{hasler2009markerless} obtain scene constraints for body pose estimation by reconstructing the scene in 3D with multiple unsynchronized moving cameras.
Some works rely on physics-inspired error terms (\eg contact and collision terms~\cite{zanfir2018monocular}), game physic engines~\cite{vondrak2012dynamical}, and scene semantic labels~\cite{savva2016pigraphs}.
Related to us, PROX~\cite{hassan2019resolving} captures person-scene interactions at a very detailed level, modelling contacts and collisions between SMPL-X body~\cite{pavlakos2019expressive} and 3D scenes.
Based on such contact and collision modelling, Zhang et al.~\cite{zhang2020place,zhang2020generating} generate human body meshes in scenes without people in a physically and semantically plausible manner.

\myparagraph{Human motion priors.} A large number of priors for smooth and natural motion have been proposed in the literature~\cite{akhter2012bilinear, arnab2019exploiting, huang2017towards, mehta2017vnect, ormoneit2001learning, rempe2020contact, ren2005data, shimada2020physcap, urtasun20063d}. 
Some priors directly apply to body joint velocity or acceleration~\cite{arnab2019exploiting,mehta2017vnect}.
Akhter et al.~\cite{akhter2012bilinear} propose a bilinear model with discrete cosine transform (DCT) basis to provide motion spatio-temporal regularity. Along this line, Huang et al.~\cite{huang2017towards} introduce a DCT prior to reconstruct body motion from multi-view input. 
Some recent work exploits physical simulation to regularize human motion. 
Shimada et al.~\cite{shimada2020physcap} assume a pre-defined virtual character as input, and fit it to monocular sequences via physics-based optimization. Rempe et al.~\cite{rempe2020contact} regress body joints and foot-ground contact from images to conduct physics-based trajectory optimization.
Kaufmann et al.~\cite{kaufmann2020infilling} design a convolutional autoencoder to infill motion of unobserved body joints and remove noise.

\myparagraph{Ours versus others.} 
In our work we design a motion smoothness prior and a motion infiller, and use them to recover realistic motions of person-scene interactions from RGB(D) videos. 
Compared to existing smoothness priors, ours is trained with high-quality AMASS sequences and the smoothness regularization is applied in the latent space. Consequently, we can produce smooth motions without degrading per-frame body pose accuracy.
Our motion infiller has a similar architecture to Kaufmann et al.~\cite{kaufmann2020infilling}, but processes body markers and predicts foot-ground contact states. 
Since body markers better constrain body DoFs, and the contact states are jointly learned with body motions, our method consistently outperforms~\cite{kaufmann2020infilling} w.r.t. motion recovery and foot skating (as shown in Sec.~\ref{sec:experiment}). Compared with~\cite{shimada2020physcap, rempe2020contact} which predict contact states by 2D joints detected from RGB images, our jointly learned contact states are better coupled with body dynamics.

%% file: sec3-method.tex
\section{Method}
\label{sec:method_v2}

\begin{figure*}
    \centering
    \includegraphics[width=\linewidth]{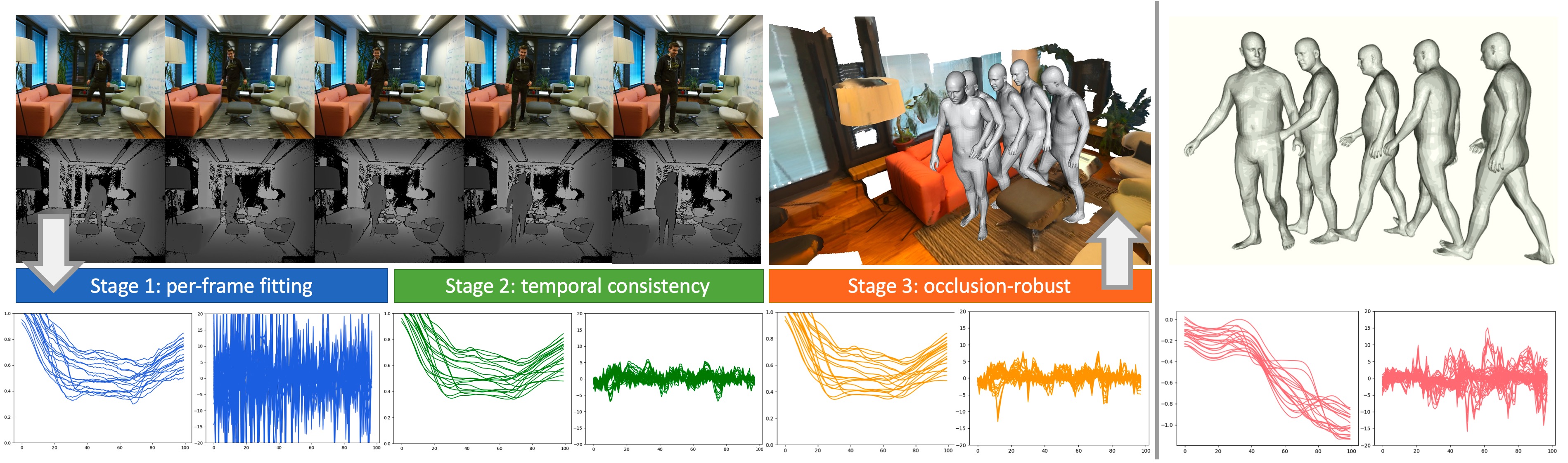}
    \caption{Illustration of our multi-stage pipeline. Provided a scene mesh and an RGBD sequence with body occlusion, our method recovers a realistic global motion, with natural person-scene interactions. 
    The markers trajectories (left) and accelerations (right) of each stage are shown at the bottom, as well as a walking sequence from AMASS~\cite{mahmood2019amass} (\textcolor{amaranth}{pink}). Note that the results from Stage~1 show large and unrealistic motion accelerations (\textcolor{blue}{blue}). The recovered motion (\textcolor{forestGreen}{green}) in Stage~2 is significantly smoother. However, it also loses the realistic accelerations (peaks in the acceleration plot) that can happen when the body interacts with the scene (e.g.~foot-to-ground contact during walking). 
    Our recovered motion from Stage~3 (\textcolor{amber}{orange}) is similar to the high-quality AMASS motion w.r.t.~both the trajectory smoothness and the acceleration patterns.
    }
    \label{fig:pipeline}
\end{figure*}

\subsection{Overview}
\label{sec: preliminaries}

We provides an overview of our approach in Fig.~\ref{fig:pipeline}. Given a sequence of RGB-D frames $\{I_t, D_t\}_{t=1}^{T}$, capturing a subject moving in a 3D scene, our goal is to reconstruct a high-quality motion, which is smooth, physically plausible, and natural.
To this end, we fit the SMPL-X parametric body model to sequence data by proceeding in three stages.

\myparagraph{SMPL-X}. 
SMPL-X~\cite{pavlakos2019expressive} represents the body as a function $\mathcal{M}_b(\boldsymbol{\gamma}, \boldsymbol{\beta}, \boldsymbol{\theta}, \boldsymbol{\phi})$, whose output is a triangle mesh with vertices $V_b \in \mathbb{R}^{10475\times 3}$. SMPL-X parameters are global translation $\boldsymbol{\gamma} \in \mathbb{R}^3$, body shape $\boldsymbol{\beta} \in \mathbb{R}^{10}$, body and hand pose $\boldsymbol{\theta}$, and facial expression $\boldsymbol{\phi} \in \mathbb{R}^{10}$.
We denote by $J(\boldsymbol{\beta})$ the 3D body skeleton joints in the neutral pose, and by $R_{\boldsymbol{\theta} \boldsymbol{\gamma}}(J(\boldsymbol{\beta})_i)$ the $i$-th joint posed according to pose $\boldsymbol{\theta}$ and translation $\boldsymbol{\gamma}$.

\myparagraph{Multi-stage pipeline.} Given the complexity of our task, we address it in a multi-stage fashion, as done in previous work~\cite{bogo2015detailed, shimada2020physcap}.
In Stage 1, we fit SMPL-X parameters to each RGB-D frame independently. This gives us a reasonable initialization, but does not ensure motion smoothness, nor deals with body-scene occlusions. We achieve temporally consistent motions in Stage 2 by introducing our smoothness prior and contact-friction term. Finally, in Stage 3 we recover plausible motions even for occluded body parts and alleviate foot skating with our motion infiller.

\subsection{Per-frame Fitting}
\label{sec:prox}
Stage~1 adopts the approach proposed in PROX~\cite{hassan2019resolving}. 
Given a RGB-D sequence, PROX fits SMPL-X to each frame separately by minimizing the objective function:
\begin{equation}
\label{eq:prox}
\begin{split}
    E_{PROX}(\boldsymbol{\gamma}, \boldsymbol{\beta}, \boldsymbol{\theta}, \boldsymbol{\phi}) &= 
    E_J + \lambda_D E_D + E_{prior} \\
    &+ \lambda_{contact} E_{contact} + \lambda_{coll} E_{coll}.
\end{split}
\end{equation}
$E_J$ penalizes the distance between the 2D joints estimated from the RGB image with OpenPose~\cite{8765346}, and the 2D projection of SMPL-X joints 
onto the image. $E_D$ penalizes the 3D distance between the human point cloud obtained from the depth frame and SMPL-X surface points visible from the camera. $E_{prior}$ combines a set of priors regularizing body pose, shape and facial expression~\cite{pavlakos2019expressive}. $E_{contact}$ encourages contact between scene vertices and a pre-defined set of body ``contact'' vertices. $E_{coll}$ penalizes scene-body interpenetration. For more details, we refer the reader to~\cite{hassan2019resolving}.

\subsection{Temporally Smooth Motion}
\label{sec:smoothness}
In Stage 2, we process the output of Stage 1. 
In order to obtain smooth and realistic motion, we design a motion smoothness prior and a physics-inspired friction term, which are then used in an optimization algorithm.

\myparagraph{Motion smoothness prior.}
Instead of enforcing smoothness explicitly on body joints as in~\cite{arnab2019exploiting, mehta2017vnect, shimada2020physcap}, we propose to learn a latent space of smooth motion. To this end, we train an autoencoder with high-quality data from AMASS~\cite{mahmood2019amass}. 

The input to our network is a sparse set of body surface markers, as in~\cite{mahmood2019amass,zhang2020we}. 
We represent the body with the locations of 81 markers (see marker placement in Supp. Mat.).
Given a sequence of $T$ frames, at each time $t$ we compute the time difference of marker locations and concatenate them to a vector of length $S$.
Then the entire sequence is represented by a 2D feature map $X_{\Delta} \in \mathbb{R}^{S\times (T-1)}$.
The network encoder $F_s$ converts $X_{\Delta}$ to its latent representation $Z$. 
Here we regard the time series $X_{\Delta}$ as an image and perform 2D convolutions as in~\cite{kaufmann2020infilling}. We do not downsample the input, so $X_{\Delta}$ and $Z$ have identical temporal resolution.
Therefore, the network captures spatio-temporal correlations with a large receptive field in the latent space, which can represent motion of overlapped body parts.
The decoder $D_s$ has a symmetric architecture with deconvolution layers. 
More details can be found in the Supp. Mat..

We train our autoencoder on the AMASS dataset with the following loss:
\begin{equation}
    \mathcal{L}_{s}(F_s, D_s) = |X_{\Delta} - X_{\Delta}^{rec}| + \alpha_s\frac{1}{S(T-2)} \sum_{t=1}^{T-2} |\boldsymbol{z}_{t+1} - \boldsymbol{z}_t |^2,
\end{equation}
where the first term is the reconstruction loss minimizing discrepancy between $X_{\Delta}$ and $X_{\Delta}^{rec}=D_s(F_s(X_{\Delta}))$, and the second term minimizes the 1st order derivative of the latent sequence $Z = F_s(X_{\Delta})=[\boldsymbol{z}_1, \boldsymbol{z}_2, ..., \boldsymbol{z}_{T-1}]$. $\alpha_s$ weights the contribution of the second term.

With a pretrained autoencoder, we design a smoothness loss to regularize motions over time.
Specifically, we take the per-frame bodies obtained from Stage 1, and concatenate their markers into a velocity map $X_{\Delta}^{opt}$. We feed this map into $F_s$, encoding it into $Z^{opt} = F_s
(X_{\Delta}^{opt})=[\boldsymbol{z}_1^{opt}, \boldsymbol{z}_2^{opt}, ..., \boldsymbol{z}_{T-1}^{opt}]$. 
The smoothness loss is given by
\begin{equation}
    E_{smooth}(\boldsymbol{\gamma}, \boldsymbol{\theta}, \boldsymbol{\phi}) = 
    \frac{1}{S(T-2)} \sum_{t=1}^{T-2} |\boldsymbol{z}_{t+1}^{opt} - \boldsymbol{z}_t^{opt} |^2.
    \label{eq:smooth}
\end{equation}
Compared to the methods working in joint space locally, our prior can better capture longer-range correlations between the motion of different body parts, hence encoding full-body dynamics. 

\myparagraph{Contact friction modelling.} 
The contact term used in Eq.~\ref{eq:prox} only considers body-scene proximity, and hence cannot prevent skating artifacts (\eg a person slides when sitting on a chair). 
To overcome this issue, we design a contact term that incorporates stationary frictions. 
Compared to methods which work with foot joints~\cite{shimada2020physcap, rempe2020contact}, our contact friction term is based on the body and scene mesh, with a more generic human-scene interaction setting, and considers also other body parts such as gluteus.

Specifically, we pre-define a set of ``contact friction'' vertices $V_c \subset V_b$, corresponding to 194 foot and 113 gluteus vertices. 
When contact occurs (\ie the distance between a body vertex to the scene mesh is smaller than 0.01m), the velocity $\boldsymbol{v}_t$ of the contacted vertex in $V_c$ is regularized: the component $\boldsymbol{v}_t$ along the scene normal $\boldsymbol{n}$ should be non-negative to prevent interpenetration, and the component tangential to the scene, $\boldsymbol{v}^{tan}_t$, should be small to prevent sliding.
Formally, this gives us:
\begin{equation}
    \boldsymbol{v}_{t} \cdot \boldsymbol{n} \geq 0, \quad |\boldsymbol{v}_t^{tan}| \leq \sigma 
    \quad \text{for} \quad t \in T_f,
\end{equation}
where $T_{f}$ is the set of frames in which vertex and scene are in contact, and $\sigma$ is a small number as a threshold.
Based on this, we formulate our contact friction term as:
\begin{equation}
    \footnotesize
    E_{fric}(\boldsymbol{\gamma}, \boldsymbol{\theta}, \boldsymbol{\phi}) = \sum_{t \in T_f}
    \left( \sum_{\boldsymbol{v}_{t}\cdot \boldsymbol{n} < 0} |\boldsymbol{v}_{t}\cdot \boldsymbol{n}|  + 
    \sum_{|\boldsymbol{v}_t^{tan}| \geq \sigma} ||\boldsymbol{v}_t^{tan}| - \sigma| \right).
    \label{eq:friction}
\end{equation}

\paragraph{Stage 2 fitting.}
We combine $E_{smooth}$ and $E_{fric}$ with the error terms in Eq.~\ref{eq:prox}, from which we remove $E_{contact}$ and $E_{D}$ to obtain a modified function $E_{PROX_M}$. We optimize the resulting objective for $N$ iterations as shown in Alg.~\ref{alg:stage2}.
\begin{algorithm}[t]
\footnotesize
\SetAlgoLined
\KwResult{Smooth motion w.r.t. SMPL-X body parameters}
 \textbf{Init}: 
 Fitted meshes from Stage 1,
 scene mesh, smoothness prior $F_s(\cdot)$\;
 \For{$i=1:N$}{
  $Z^{opt} = F_s(X_{\Delta}^{opt})$\;
  compute $E_{smooth}$ with Eq.~\eqref{eq:smooth}\;
  compute $E_{fric}$ with Eq.~\eqref{eq:friction}\;
  minimize $E_{PROX_M} + E_{smooth} + E_{fric}$  
 }
 \caption{smooth motion recovery in Stage 2.}
 \label{alg:stage2}
\end{algorithm}

\subsection{Recovering Motion under Occlusion}
Although Alg.~\ref{alg:stage2} produces smooth motions, it cannot recover realistic body motion under occlusion, which frequently happens in person-scene interactions.
Therefore, we design a motion infilling network, train it on AMASS, and exploit it as a prior in an optimization algorithm.

\begin{figure}[t]
    \centering
    \includegraphics[width=\linewidth]{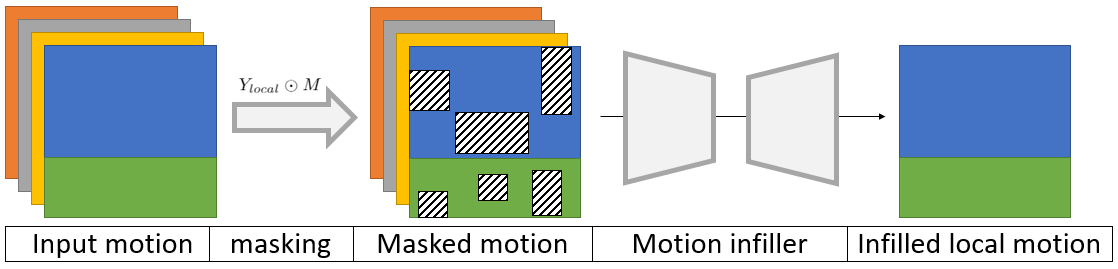}
    \caption{Illustration of our motion infilling network. The blue and green color denote the marker local coordinate and the foot-ground contact state, respectively. Yellow, gray and orange color represent the root translational velocity $[t_1, t_2]$ and rotational velocity $\gamma$, respectively. Note that masking is only applied to the local motion. The motion infiller takes global motion as input, and predicts local motion. }
    \label{fig:motion_infiller}
\end{figure}

\myparagraph{Motion infilling network.}  Figure~\ref{fig:motion_infiller} shows an overview of our convolutional infilling network.
Unlike other motion infilling models e.g.~\cite{kaufmann2020infilling}, our infiller takes body surface markers as input, and infers motions and contact states jointly.

Here we represent markers in a local coordinate system as in~\cite{holden2016deep,kaufmann2020infilling}: for each frame $t$, marker locations are relative to the position of the body root, which is the pelvis projected to the ground.
In addition, the body global configuration is represented by the root's translational velocity $\boldsymbol{t}\in \mathbb{R}^{2}$ and rotational velocity $\gamma \in \mathbb{R}$ around the up-axis. 
Moreover, we select two markers per foot and check whether they contact with the ground at each frame. The marker is deemed in contact with the ground if its velocity is smaller than 20cm/s and its height above the ground is lower than 10cm.
Finally, we arrange the motion sequence into a 3D tensor $Y \in \mathbb{R}^{P\times T \times 4}$ with 4 channels. In the first channel $Y_{local}$, each column denotes a vector concatenating local body marker positions and contact labels of this frame, and $P$ is the vector dimension.
The last three channels $Y_{root}$ consist of the repeated entries of global trajectory velocities ${t}_1$, ${t}_2$ and $\gamma$ respectively,
which allows us to couple global and local motion more closely than~\cite{holden2016deep,kaufmann2020infilling}.

During training, we set a spatio-temporal visibility mask $M \in \{0,1\}^{P \times T}$ (1 denotes visible, and 0 otherwise) to corrupt the local motion with $\Tilde{Y}_{local} = Y_{local} \odot M$, 
where $\odot$ denotes element-wise matrix multiplication. 
Since most (upper) body parts are often visible in practice and it is easy to estimate the root's motion, we assume the global trajectory is given, and only mask local pose. To generate plausible occlusion masks for training on AMASS, we sample masks computed from the PROX dataset~\cite{hassan2019resolving}, where we leverage depth to reason about occlusions. Note that contact labels are masked whenever feet are not visible.
With the masked motion $\Tilde{Y}=[\Tilde{Y}_{local}, Y_{root}]$, we train the infiller autoencoder $G$ to reconstruct the full local motion by minimizing:
\begin{equation}
    \mathcal{L}_{infill}(G) = h(G(\Tilde{Y}), Y_{local}),
\end{equation}
where $h(\cdot)$ is the L1 loss for local marker coordinates, and the binary cross entropy (BCE) loss for the contact labels.

\myparagraph{Per-instance self-supervised learning.}
\label{sec:pft}
To better leverage visible body parts during testing, we fine-tune the pre-trained motion infiller on each individual test motion sequence to adapt the learned general prior to per-instance. Unlike~\cite{joo2020exemplar}, our fine-tuning procedure is self-supervised. Specifically, given a partially occluded test sequence $Y$, and occluded markers described by mask $M$, we fine-tune the network parameters by exploiting the visible markers in the sequence via minimizing
\begin{equation}
    \mathcal{L}_{finetune}(G) = h(G(\Tilde{Y}), Y_{local}) \odot M.
    \label{eq:fine-tune}
\end{equation}

We show in Sec.~\ref{sec:experiment} that this per-instance self-supervised learning effectively increases prediction accuracy for both visible and invisible body parts.

\paragraph{Stage 3 fitting~(Alg.~\ref{alg:stage3}).}
Given results from Stage 2, we combine the global configurations and local markers produced by the fine-tuned infiller, and reconstruct marker global locations $\hat{X}$ and foot contact labels $\hat{C}$. We define an error term as
\begin{equation}
\label{eq:infill}
    \begin{aligned}
    E_{infill} (\boldsymbol{\gamma}, \boldsymbol{\theta}, \boldsymbol{\phi}) &= |\hat{X} - X^{opt}| \odot (1-M_b)  \\
    & + \sum_{t=1}^T \sum_{k \in K} \hat{c}_k^t \cdot d(v_k^t, a),
    \end{aligned}
\end{equation}
where $X^{opt}$ is the marker global location from the SMPL-X body to optimize, $M_b$ is the occlusion mask for the body, and $K$ is the set of foot vertices.
For foot vertex $k$, $\hat{c}_k^t = 1$ if its nearest foot marker contact label is $1$, $0$ otherwise, and $v_k^t$ is the absolute magnitude of velocity at frame $t$.
$d(v_k^t, a)$ corresponds to $|v_k^t - a| $ if $v_k^t \geq a$, $0$ otherwise.
We set the foot velocity threshold $a$ as 10cm/s.

\begin{algorithm}[t]
\footnotesize
\SetAlgoLined
\KwResult{Infilled motion in the presence of occlusions}
 \textbf{Init}: Results from Stage 2, scene mesh, smoothness prior $F_s(\cdot)$, motion infiller $G(\cdot)$\;
 \textbf{Step 1}: fine-tune $G(\cdot)$ according to Eq.~\eqref{eq:fine-tune}\;
 \textbf{Step 2}: compute $\hat{X}$, $\hat{C}$ from $G(\cdot)$\;
 \textbf{Step 3}: the optimization loop\;
     \For{$i=1:N$}{
      $Z^{opt} = F_s(X_{\Delta}^{opt})$\;
      compute $E_{smooth}$ with Eq.~\eqref{eq:smooth}\;
      compute $E_{fric}$ with Eq.~\eqref{eq:friction}\;
      compute $E_{infill}$ with Eq.~\eqref{eq:infill}\;
      minimize $E_{PROX_M} + E_{smooth} + E_{fric} + E_{infill}$
     }
\caption{occluded motion recovery in Stage 3.}
\label{alg:stage3}
\end{algorithm}

%% file: sec4-experiment.tex
\section{Experiments}
\label{sec:experiment}

\subsection{Datasets}

\myparagraph{AMASS~\cite{mahmood2019amass}}.
\textbf{AMASS} collects 15 high-quality mocap datasets, with $11263$ motions from $344$ subjects. For each sequence, \textbf{AMASS} provides per-frame SMPL-H~\cite{MANO:SIGGRAPHASIA:2017} parameters obtained via MoSh++ (\ie fitting SMPL-H to mocap markers).
We downsample the sequences to 30fps, and trim them to clips of 120 frames for training. 
Similar to~\cite{zhang2020we}, for each clip we reset the world coordinate to the pelvis joint in the first frame. The $x$-axis is the horizontal component of the direction from the left hip to right hip, the $y$-axis points forward, and the $z$-axis points upward.
We exclude TCD\_handMocap, TotalCapture, SFU, SSM\_synced, KIT and EKUT, and use the rest to train our motion smoothness and infilling models. 
We exclude {TCD\_handMocap}, {TotalCapture}, {SFU} from training since we use them to evaluate our motion infilling method. 
We do not use EKUT, KIT and SSM\_synced due to their inconsistent frame rate.

\myparagraph{PROX~\cite{hassan2019resolving}}.
We use this dataset to test our models and optimization algorithms in Stage 2 and Stage 3. 
{\bf PROX} collects monocular RGB-D sequences from 20 subjects moving in and interacting with 12 different indoor scenes.
A Kinect-One sensor~\cite{kinect1} is employed to capture the sequences at 30fps, and the 3D reconstructions of the static scenes are provided. SMPL-X parameters are fitted to RGB-D data in each frame (see Sec.~\ref{sec:prox}) to reconstruct 3D bodies.
Following the same pre-processing procedure for {\bf AMASS}, we trim the sequences, reset the pelvis coordinate, and obtain 205 clips of 100 frames each for evaluation.

\myparagraph{3DPW~\cite{von2018recovering}}. As with {\bf PROX}, we use this dataset for evaluation.
{\bf 3DPW} fits SMPL to IMUs and RGB videos, mostly captured in in-the-wild scenario.
Although the provided per-frame SMPL fits are accurate, the motion across frames has jitters and temporal discontinuities. 
As above, we pre-process the motion sequences and obtain 300 clips of 100 frames for evaluation. 
Since sequences are captured with a moving camera, global SMPL body configurations are not reconstructed accurately. Hence, for {\bf 3DPW} we test our priors applying them only on \emph{local} motions. 
Namely, the body pelvis joints in different frames are aligned, and joint positions are defined with respect to the local coordinate system of each individual frame.

\subsection{Evaluation of Motion Smoothness Prior}
We compare our motion smoothness prior (denoted by `Ours-SP') against three optimization-based baselines: 
the DCT-based prior from~\cite{huang2017towards}; minimizing velocity magnitude (L2-V)~\cite{li2019estimating, tripathi2020posenet3d, arnab2019exploiting, zhou2016sparseness}; minimizing acceleration magnitude (L2-A)~\cite{li2019estimating, mehta2017vnect, shimada2020physcap}.
For all methods, we combine them with $E_J, E_{prior}$ in Eq.~\ref{eq:prox} and minimize the resulting objective function to fit SMPL-X to data.
Specifically, the objective function of Ours-SP consists of $E_J, E_{prior}$ and $E_{smooth}$ in Eq.~\ref{eq:smooth}.
We evaluate the fits on both \textbf{PROX} and \textbf{3DPW}.

\subsubsection{Metrics}
\myparagraph{2D joint accuracy.} 
This metric is only used for {\bf PROX}.
We manually annotate 2D body joints on 542 frames via Amazon Mechanical Turk (AMT). The AMT annotations are in the OpenPose~\cite{8765346} coco-25 format (including 25 body joints), and are converted to the SMPL-X body joint format for evaluation. Following~\cite{pavlakos2019expressive}, the neck, left and right hip joints are excluded from evaluation due to their definition ambiguity. We report the average L2 norm of 2D joint errors (2DJE) between our results and annotations. 

\myparagraph{3D accuracy.} 
This metric is only used for {\bf 3DPW}.
Following~\cite{von2018recovering}, we report the mean per joint position error (MPJPE) and per vertex error (PVE) with aligned body pelvis between our estimated motions and the ones provided by \textbf{3DPW}.
We expect that an effective motion smooth prior can improve motion temporal consistency while preserving the original body configuration quality.
Therefore, the lower these two scores are, the better. 
However, for an exhaustive evaluation, 3D accuracy should be combined with a metric assessing motion smoothness.

\myparagraph{Motion smoothness.} Ideally, recovered motions should resemble real ones as much as possible.
Translating this into a metric, we use the Power Spectrum KL divergence (PSKL)~\cite{hernandez2019human} to
measure the distribution distance between our results and {\bf AMASS} motion sequences. Specifically, we evaluate PSKL w.r.t. the acceleration distribution for both body markers and SMPL-X joints on \textbf{PROX}, and for SMPL joints on \textbf{3DPW}.
Since PSKL is not a symmetric measure, we report the numbers for both directions.
Smaller values of PSKL indicate better performance (see Supp. Mat. for more details).

\myparagraph{Human-scene interpenetration.} 
We assess the degree of human-scene interpenetration on \textbf{PROX} by using the non-collision score adopted in~\cite{zhang2020place,zhang2020generating}. It measures the ratio between the number of body vertices with non-negative scene SDF values, and the total number of body vertices, \ie, the ratio of body vertices that do not interpenetrate with the scene mesh. We report the average non-collision scores over all frames, and denote it as `NonColl'. A higher value indicates fewer human-scene interpenetration.

\vspace{-0.1cm}

\subsubsection{Results}
Tab.~\ref{tab:eval-prox} and Tab.~\ref{tab:eval-3dpw} show the results of motion smoothness evaluation on \textbf{PROX} and \textbf{3DPW}, respectively. 
For both datasets, the originally provided motions have the largest PSKL score measured w.r.t {\bf AMASS}, indicating that motions are not natural. 
Compared to all baselines, our method achieves the lowest PSKL scores in both directions, suggesting that it produces more natural motions.
On \textbf{PROX}, all methods achieve comparable non-collision scores.
Our method achieves a lower 2D pose error compared to the original \textbf{PROX} data and baseline methods. 
On \textbf{3DPW}, our method has small MPJPE/PVE while reporting the best PSKL scores. 
These results demonstrate that our method can generalize well over different datasets for both global motions and local motions.

\begin{figure}
    \centering
    \includegraphics[width=\linewidth]{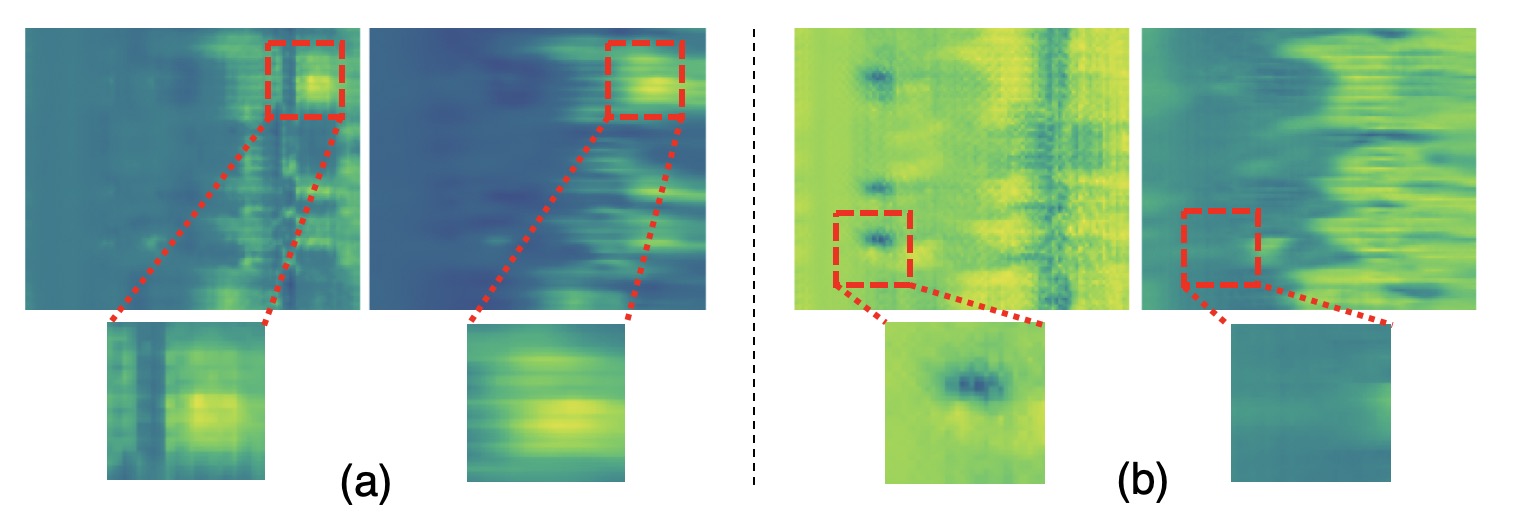}
    \caption{Illustration of two channels ((a) and (b)) of our motion smoothness model latent sequences $Z$. In each sub-figure, the left and right plots show the result before and after Alg.~\ref{alg:stage2}, respectively. The row and column in each plot denote feature dimension and time, respectively.
    }
    \label{fig:latent_z}
\end{figure}

Overall, our method consistently outperforms other baselines, by significantly improving motion naturalness while preserving per-frame pose accuracy.
This is due to the fact that we learn our smoothness prior from the rich and diverse {\bf AMASS} data, and apply regularization in latent space.
In contrast, baseline methods only encourage motion smoothness of disjoint local body parts, and hence have larger gaps to high-quality AMASS motions. 
Fig.~\ref{fig:latent_z} shows examples of latent sequences obtained from {\bf PROX} sequences. After fitting using our prior, latent sequences become smoother along the time axis and jitters are removed.

\begin{table}[t!]
\centering
\scriptsize
\caption{\textbf{Evaluation of motion smoothness and infilling priors on PROX.} 
PSKL-M and PSKL-J denote PSKL computed on markers and joints, respectively. (P,~A) denotes PSKL(PROX,~AMASS), and (A,~P) the reverse direction. For each metric, the best result is in boldface.}
\begin{tabular}{lcccccc}
  \toprule[1pt]

  & 2DJE $\downarrow$ & \multicolumn{2}{c}{PSKL-M $\downarrow$} & \multicolumn{2}{c}{PSKL-J $\downarrow$} & NonColl $\uparrow$ \\
  \cmidrule(lr){3-4}   \cmidrule(lr){5-6} 

  Methods & & (P,A) & (A,P) & (P,A) & (A,P) & \\  

  \midrule
  PROX~\cite{hassan2019resolving} & 20.94 & 1.439 & 2.441 & 1.464 & 2.491 & \textbf{0.955} \\
  DCT~\cite{huang2017towards} & 20.96 & 0.847 & 1.083 & 0.937 & 1.169 & \textbf{0.955} \\
  L2-A & 21.68 & 0.429 & 0.396 & 0.481 & 0.441 & \textbf{0.955} \\
  L2-V & 21.65 & 0.551 & 0.525 & 0.571 & 0.536 & 0.954 \\
  Ours-SP & \textbf{20.64} & \textbf{0.249} & \textbf{0.256} & \textbf{0.272} & \textbf{0.275} & 0.954 \\
  
  \midrule
  Ours-S2 & 20.40 & 0.273 & 0.255 & 0.297 & 0.275 & 0.977 \\
  Ours-S3 & \textbf{20.23} & \textbf{0.236} & \textbf{0.234} & \textbf{0.256} & \textbf{0.255} & \textbf{0.979} \\

  \bottomrule[1pt]
 \end{tabular}
\label{tab:eval-prox}
\end{table}

\begin{table}[t!]
\centering
\footnotesize
\caption{\textbf{Evaluation of our motion smoothness prior on 3DPW.} 
PSKL-J denotes PSKL of joints. (3D,~A) denotes PSKL(3DPW,~AMASS), and (A,~3D) the reverse direction. For each metric, the best result is in boldface.}
\begin{tabular}{lcccc}
  \toprule[1pt]

  & MPJPE$\downarrow$  & PVE$\downarrow$  & \multicolumn{2}{c}{PSKL-J $\downarrow$}  \\
  \cmidrule(lr){4-5}  
  Methods & & & (3D, A) & (A, 3D) \\  
  \midrule
  3DPW~\cite{von2018recovering} & - & - & 0.348 & 0.376 \\
  DCT~\cite{huang2017towards} & \textbf{0.005} & \textbf{0.007} & 0.242 & 0.273 \\
  L2-A & 0.006 & 0.009 & 0.177 & 0.204 \\
  L2-V & 0.019 & 0.025 & 0.257 & 0.271 \\
  Ours-SP & \textbf{0.005} & 0.008 & \textbf{0.173} & \textbf{0.197} \\
  
  \bottomrule[1pt]
 \end{tabular}
\label{tab:eval-3dpw}
\end{table}

\begin{table*}[]
\centering
\footnotesize
\caption{\textbf{Evaluation for motion infilling prior on AMASS.}
MPJPE-L / MPMPE-L denotes MPJPE / MPMPE for the masked lower body part. Finetune denotes per-instance self-supervised learning. For each metric, the best result is in boldface.}
\begin{tabular}{llcccccc}
  \toprule[1pt]
  &Methods & MPJPE $\downarrow$ & MPMPE $\downarrow$ & VPE $\downarrow$ & MPJPE-L $\downarrow$ & MPMPE-L $\downarrow$ & Foot Skating $\downarrow$ \\
  
  \midrule
  \multirow{2}*{Ours vs baseline} 
  &Kaufmann et al.~\cite{kaufmann2020infilling} & 0.022 & 0.026 & 0.025 & 0.037 & 0.036 & 0.237 \\ 
  &Ours-IP & \textbf{0.014} & \textbf{0.016} & \textbf{0.012} & \textbf{0.034} & \textbf{0.033} & \textbf{0.182} \\
  
  \midrule
  
  \multirow{5}*{Ablation study}
  &Ours-IP w/o Opt w/o finetune & - & 0.025 & - & - & 0.040 & - \\
  &Ours-IP w/o Opt & - & \textbf{0.015} & - & - & \textbf{0.036} & - \\

\cmidrule(lr){2-8}

  &Ours-IP w/o finetune heuristic contact & 0.020 & 0.024 & 0.021 & 0.040 & 0.038 & 0.257 \\
  &Ours-IP w/o finetune & 0.020 & 0.023 & 0.021 & 0.038 & 0.036 & \textbf{0.178} \\
  &Ours-IP heuristic contact & \textbf{0.014} & \textbf{0.017} & \textbf{0.013} & \textbf{0.036} & \textbf{0.035} & 0.265 \\
  \bottomrule[1pt]
 \end{tabular}
\label{tab:eval-infill}
\end{table*}

\subsection{Evaluation of Motion Infilling Prior}
We compare our proposed motion infilling prior (denoted by `Ours-IP') with the infiller from Kaufmann et al.~\cite{kaufmann2020infilling} on \textbf{AMASS}. On top of these networks, we additionally fit SMPL-X parameters to the infilled markers/joints, so as to perform fair comparison.
The fitting function is:
\begin{equation}
\label{eq:infill_amass_temporal}
    E_{amass}  = E_{3D} + E_{prior} + E_{smooth} + E_{foot},
\end{equation}
where $E_{3D}$ is the error between infilled marker positions and the corresponding markers on the SMPL-X body to optimize, and $E_{prior}$ is the prior term for body and hand pose.
For our method, $E_{foot}$ is the second term in Eq.~\ref{eq:infill}.
For the baseline method, we define $E_{foot}$ using a heuristic: foot-ground contact happens if the foot marker distance from the ground is smaller than 10cm (see Supp. Mat.).

We randomly select 130 sequences from our {\bf AMASS} test set, in order to remove redundant motions and reduce computational cost.
To simulate the occlusions occurring in real person-scene interactions and absent in {\bf AMASS}, at evaluation time, for the network input of both methods, we mask out all markers belonging the lower part of the body and the contact labels in all frames. 
Furthermore, we evaluate the proposed motion infilling prior on \textbf{PROX} in terms of 2D joint accuracy, PSKL and non-collision score.

\subsubsection{Metrics}
\myparagraph{3D accuracy.}
We report the mean position error for joints (MPJPE), body markers (MPMPE) and body vertices (PVE) in \textit{global} coordinates between the infilled motions and the motions from \textbf{AMASS}. 
We compute these three metrics for the full body, and also compute MPJPE and MPMPE for the masked body parts.

\myparagraph{Foot skating.} 
Following~\cite{zhang2020we}, we adopt the ``foot skating ratio'' as another measure of motion naturalness.
We compute it by considering the two markers located on the left and right foot heels. 
We define skating as happening when the velocity of both foot markers exceeds 10cm/s and their height above the ground is lower than 10cm.

\subsubsection{Results}
The results on \textbf{AMASS} are shown in Tab.~\ref{tab:eval-infill}. 
Our infiller consistently outperforms the baseline for all metrics.
In particular, our model reconstructs more accurate motions with all the three representations (body marker, joint and vertex). Besides, we obtain smaller reconstruction errors for the lower part of the body (MPJPE-L and MPMPE-L).

Compared to the heuristic foot-ground contact rule used for the baseline, our predicted contact labels alleviate foot skating more effectively, and recover foot dynamics during optimization. This is also verified in the ablation study (Ours-IP heuristic contact), where we replace the predicted contact labels by the same heuristic contact rule used in the baseline. A probable reason is that our model learns foot-ground contact and whole body motion jointly, and hence can predict the two more consistently.

In addition, the ablation study suggests that our model performance is consistently improved by the self-supervised fine-tuning (see Sec.~\ref{sec:pft}), before SMPL-X fitting (Ours w/o Opt) and after, for both whole body and occluded parts. 
This indicates that our motion infiller effectively adapts itself to test instances, exploiting more useful information from the unmasked body parts of the input. 

The last two rows in Tab.~\ref{tab:eval-prox} show results of the motion infilling prior on \textbf{PROX}. Compared with results of Stage 2 without motion infilling (Ours-S2), Stage 3 (Ours-S3) has an acceleration closer to \textbf{AMASS}, and lower 2D joint errors. 
Finally, to assess the stages of our pipeline, we swap Stage 2 and Stage 3, and find that the motion infiller works poorly when taking jittered \textbf{PROX} data as input (see Supp. Mat.). Also, the model overfits to the noisy input when performing the self-supervised test fine-tuning.

%% file: sec5-conclusion.tex
\section{Conclusion}
\label{sec:conclusion}
In this paper, we propose a novel motion smoothness prior and a contact-aware motion infilling prior learned from high-quality motion capture data, which effectively learn intrinsic full-body dynamics of smooth motions and recover body parts occluded from the camera view. On top of that, we introduce a new multi-stage optimization pipeline which incorporates the motion priors and a physics-inspired contact friction term, and reconstructs smooth, accurate and occlusion-robust global motions with physically plausible human-scene interactions in complex 3D environments. 
Nevertheless, there are limitations in the current approach. For instance, human movement is rooted in physics. The current pipeline only incorporates intuitive physics terms (\eg contact, interpenetration and friction); it is a very promising and challenging research direction to employ more physics inspired motion modeling, in combination with the powerful data-driven motion priors.

{\small
\myparagraph{Acknowledgements.}
This work was supported by the Microsoft Mixed Reality \& AI Zürich Lab PhD scholarship. We sincerely thank Shaofei Wang and Jiahao Wang for proofreading.
}

%% file: sec6-appendix.tex
\begingroup
\onecolumn 

\appendix
\begin{center}
\Large{\bf Learning Motion Priors for 4D Human Body Capture in 3D Scenes \\ **Appendix**}
\end{center}

\setcounter{page}{1}
\setcounter{table}{0}
\setcounter{figure}{0}
\renewcommand{\thetable}{S\arabic{table}}
\renewcommand\thefigure{S\arabic{figure}}

\section{Architecture Details}
The model architecture for motion priors is illustrated in Fig.~\ref{fig:appendix-architecture}. 
The motion smoothness prior and the motion infilling prior share a similar network architecture,
The encoder includes 5 consecutive convolution blocks, with each block containing [conv3x3, LeakyReLU, conv3x3, LeakyReLU, MaxPooling] layers.
The motion smoothness prior has the feature channel of [32, 64, 64, 64, 64] for the output of each encoder block. The motion infilling prior has the feature channel of [32, 64, 128, 256, 256] for the output of each encoder block.
The decoder includes 5 deconvolution blocks, with each block containing [deconv3x3, LeakyReLU, deconv3x3, LeakyReLU] layers.
For the motion smoothness prior, since the smooth constraint (Eq.~\ref{eq:smooth}) works on the latent space, we do not downsample the features so that the latent space can preserve the full spatial-temporal resolution the same as the input motion, to model smooth full-body dynamics without losing motion details, thus the  MaxPooling layer is not included in the motion smoothness prior.

\begin{figure}[h!]
    \centering
    \includegraphics[width=0.95\linewidth]{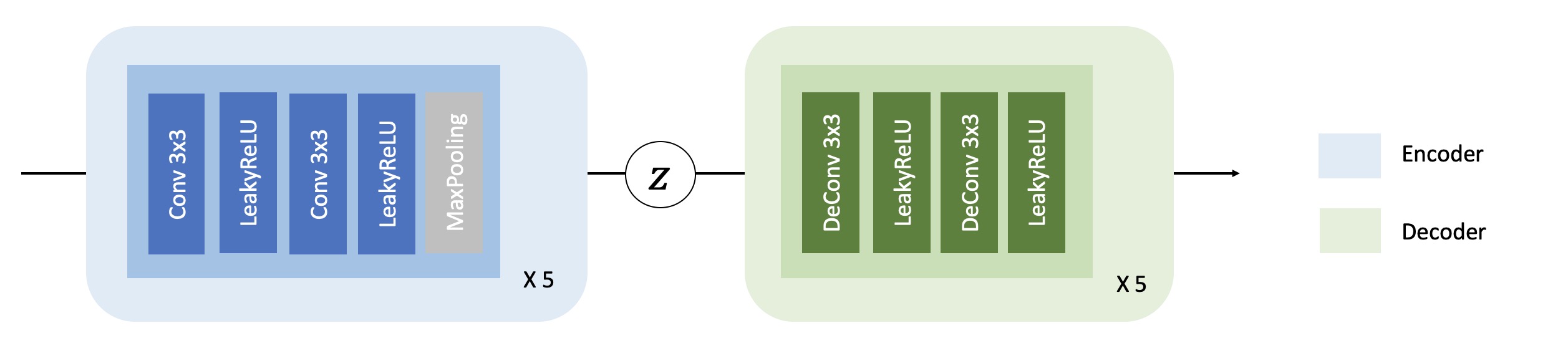}
    \caption{Model architecture for motion priors. The encoder includes 5 consecutive [conv3x3, LeakyReLU, conv3x3, LeakyReLU, MaxPooling] blocks, and the decoder includes 5 consecutive [deconv3x3, LeakyReLU, deconv3x3, LeakyReLU] blocks. The MaxPooling layers are only included in the motion infilling prior network.
    }
    \label{fig:appendix-architecture}
\end{figure}

\section{Experiment Details}
\subsection{Implementation Details}
The proposed algorithm is implemented with PyTorch 1.4.0. We use a single TITAN RTX GPU for training and optimization.
For the motion smoothness prior and motion infilling prior training, we use ADAM as the optimizer ($\beta_1=0.9$, $\beta_2=0.999$) with the learning rate 1e-4. The motion smoothness prior is trained for 150 epochs with a batch size of 60, and the motion infilling prior is trained for 900 epochs with a batch size of 120. 
For the proposed multi-stage optimization pipeline on PROX~\cite{hassan2019resolving}, we use ADAM as the optimizer ($\beta_1=0.9$, $\beta_2=0.999$) with the learning rate 5e-3.  In both Stage 2 and Stage 3, we optimize for 900 steps for each motion clip of 100 frames.

\subsection{Marker Placement}
The body surface marker placement is illustrated in Fig.~\ref{fig:appendix-markers}. We choose 67 markers on the SMPL-X body mesh surface, following the SSM2 marker setting in~\cite{mahmood2019amass}. Furthermore, we select additional 14 markers on the face and fingers to enforce smoothness over hand motions and facial expressions. Note that the additional 14 markers are only utilized in the motion smoothness prior, as we mainly focus on motion infilling for lower part of the body in the motion infilling prior.
Compared with body joints, body surface markers can better model degrees-of-freedom (DoFs) and incorporate body shape information~\cite{zhang2020we}.
\begin{figure}[h!]
    \centering
    \includegraphics[width=0.95\linewidth]{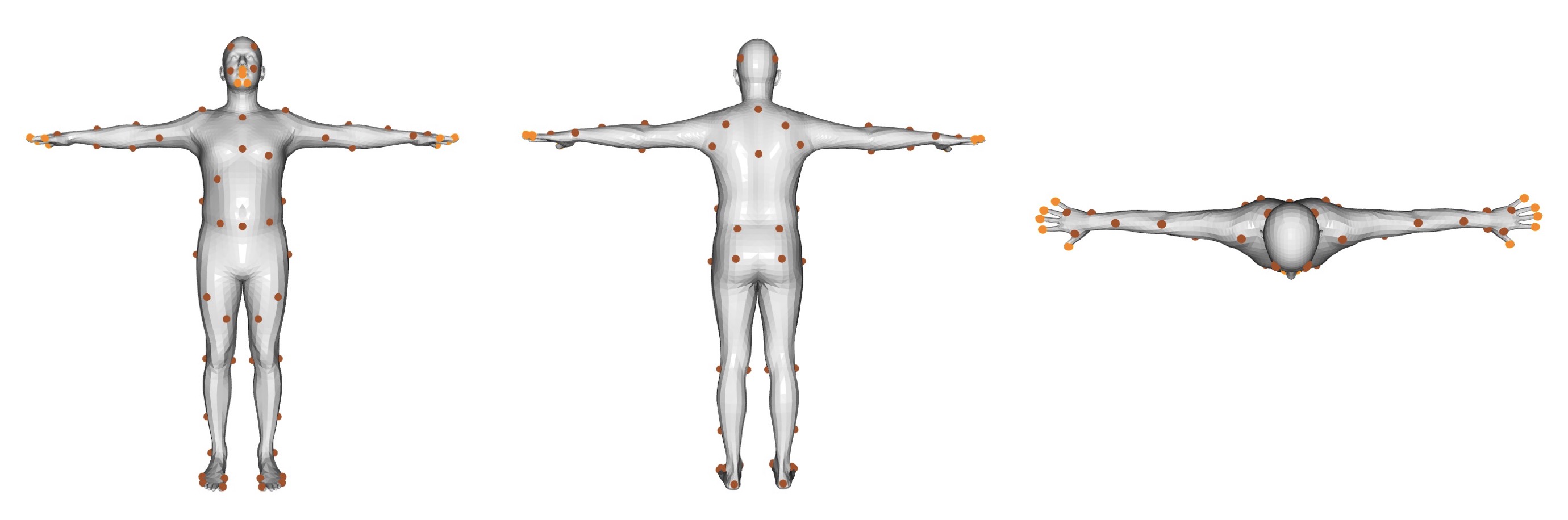}
    \caption{Marker placement for motion priors. The body markers are denoted by spheres over the SMPL-X body surface, following the marker setting (brown) in~\cite{mahmood2019amass}, with 14 additional markers (orange) on the face and fingers. From left to right: the front view, the back view and the top view.
    }
    \label{fig:appendix-markers}
\end{figure}

\subsection{Evaluation Details}
\paragraph{Power spectrum KL divergence (PSKL).}
We use PSKL to measure the distribution distance between two datasets, as in~\cite{hernandez2019human}. Formally, given a motion sequence of $T$ frames, and each frame represented by $F$ features, the power spectrum of each feature sequence $s_f$ is $PS(s_f) = ||FFT(s_f)||^2$. $x,y,z$ accelerations of each frame are used as the features and $F=3M$, where $M$ denotes the number of body markers or joints. 
The average power spectrum for feature $f$ over $N$ motion sequences on a dataset $C$ is computed as: 
\begin{equation}
    PS(C|f) = \frac{1}{N} \sum_{n=1}^N PS(s_f),
\end{equation}
and $PS(C|f)$ is normalized along the frequency dimension. PSKL between datasets $C$ and $D$ is the average power spectrum KL divergence over all feature dimensions:
\begin{equation}
    PSKL(C, D) = \frac{1}{F} \sum_{f=1}^F \sum_{e=1}^E ||PS(C|f)|| \ast log (\frac{||PS(C|f)||}{||PS(D|f)||)},
\end{equation}
where $e$ is frequency. KL divergence is asymmetric, thus both directions PSKL(C,D) and PSKL(D,C) are computed.

\paragraph{Motion infilling prior experiments.}
Here we describe the fitting procedure for the motion infilling prior evaluation on AMASS~\cite{mahmood2019amass} in detail. 
For both our proposed method and the baseline method, firstly we implement per-frame fitting with the following objective function:
\begin{equation}
    E_{PF}  = E_{3D} + E_{prior},
\end{equation}
where $E_{3D}$ is the L1 loss between infilled marker positions inferred by the infilling network and marker positions of the SMPL-X body to optimize, and $E_{prior}$ is the prior term for body pose and hand pose. The per-frame fitting aims to provide a good initialization for the temporal fitting. Initialized from per-frame fitting results, the temporal fitting is implemented by minimizing:
\begin{equation}
    E_{TF}  = E_{3D} + E_{prior} + E_{smooth} + E_{foot},
\end{equation}
where $E_{smooth}$ is the proposed smooth prior term in Eq.~\ref{eq:smooth}. For our proposed method, $E_{foot}$ is the second term in Eq.~\ref{eq:infill}, which penalizes foot vertex velocity according to the predicted foot-ground contact states. For the baseline method, $E_{foot}$ is defined by a heuristic cue:
\begin{equation}
    E_{foot} = \sum_{k, t: z_{k}^t \leq z_{thres}} d(v_k^t, a),
\end{equation}
where $z_{k}^t$ is the distance from the ground of foot joint $k$ at frame $t$, with $z_{thres}$ set to 10cm. $v_k^t$ is the absolute velocity magnitude of foot joint $k$ at frame $t$. 
$d(v_k^t, a)$ corresponds to $|v_k^t - a| $ if $v_k^t \geq a$, $0$ otherwise. 
Foot velocity threshold $a$ is set to 10cm/s. This term penalizes foot joint velocity when its distance to the ground is smaller than 10cm.

\begin{table}[t!]
\centering
\caption{\textbf{Ablation study for swapping the proposed Stage 2 and Stage 3 on PROX.} 
PSKL-M and PSKL-J denote PSKL computed on markers and joints, respectively. (P,~A) denotes PSKL(PROX,~AMASS), and (A,~P) the reverse direction. For each metric, the better result is in boldface.}
\begin{tabular}{lcccccc}
  \toprule[1pt]

  & 2DJE $\downarrow$ & \multicolumn{2}{c}{PSKL-M $\downarrow$} & \multicolumn{2}{c}{PSKL-J $\downarrow$} & NonColl $\uparrow$ \\
  \cmidrule(lr){3-4}   \cmidrule(lr){5-6} 

  Methods & & (P,A) & (A,P) & (P,A) & (A,P) & \\  

  \midrule
  Ours-S3 & \textbf{20.23} & \textbf{0.236} & \textbf{0.234} & \textbf{0.256} & 0.255 & \textbf{0.979} \\
  Ours-S3-S2 & 20.64 & 0.273 & 0.236 & 0.307 & \textbf{0.254} & 0.972 \\

  \bottomrule[1pt]
 \end{tabular}
\label{tab:appendix-swap-S2-S3}
\end{table}

\paragraph{Swap Stage 2 \& Stage 3.}
As the motion infilling prior is trained with high-quality data on AMASS, and the proposed self-supervised test fine-tuning relies on the motion of visible body parts, it requires smooth motions as input for good performance. Therefore we first recover temporal consistent motion by the Stage 2, and then include the motion infilling prior in Stage 3. 
As shown in Tab.~\ref{tab:appendix-swap-S2-S3}, if we swap Stage 2 and Stage 3 (denoted by `Ours-S3-S2'), the overall motion naturalness (PSKL score) will degrade, as well as the pose accuracy (2DJE).

\section{Comparison with Regression-based Methods}
On PROX, we additionally compare the proposed motion smoothness prior (Ours-SP) with three regression-based denoising methods~\cite{holden2016deep, kim2019human, wang2019spatio}. These methods directly output smooth motions represented by body joints by taking noisy motion as input. For a fair comparison, we train them on AMASS, adding Gaussian noise to the input motion. 

Tab.~\ref{tab:regressor-baselines} shows that our smoothness prior (Ours-SP) achieves significantly higher joint accuracy, and produces more realistic motions according to the PSKL scores.
The regression-based methods are trained with synthesized noise, which limits their generalizability to different noise distributions, and frequently produces inaccurate global reconstruction, while our motion prior is trained with clean motions, and works very well both on 3DPW captured by IMU sensors and PROX captured by Kinect. 
Besides, it is unclear how to incorporate the 3D scene constraints directly into the regressors. In contrast, our motion priors and human-scene interaction constraints can be unified in an optimization framework to produce realistic motions that satisfy 3D scene constraints.

\begin{table}[t!]
\centering
\caption{\textbf{Comparison with regression-based denoising models on PROX.} 2DJE denotes the 2D joint accuracy. PSKL-J denotes PSKL of joints. (P, A) denotes PSKL(PROX, AMASS), and (A, P) the reverse direction. For each metric, the best result is in boldface.} 
\begin{tabular}{lccc}
\toprule[1pt]
  Methods & 2DJE $\downarrow$ & PSKL-J (P,A) $\downarrow$ & PSKL-J (A,P) $\downarrow$ \\
\midrule
Wang et al.~\cite{wang2019spatio} & 140.47 & 0.294 & 0.303  \\
Holden et al.~\cite{holden2016deep} & 62.97 & 0.487 & 0.462  \\
Kim et al.~\cite{kim2019human} & 66.05 & 0.285 & 0.278 \\
Ours-SP & \textbf{20.64} & \textbf{0.272} & \textbf{0.275}  \\
 \bottomrule[1pt]
 \end{tabular}
\label{tab:regressor-baselines}
\end{table}